\documentclass[runningheads]{llncs}
\usepackage{graphicx}

\usepackage{url}
\usepackage{subcaption}
\usepackage{tikz}
\usepackage{comment}
\usepackage{amsmath,amssymb} 
\usepackage{color}

\usepackage{cellspace, multirow, booktabs}
\usepackage[accsupp]{axessibility}  

\begin{document}
\pagestyle{headings}
\mainmatter
\def\ECCVSubNumber{8}  

\title{MaRF: Representing Mars as Neural Radiance Fields} 

\titlerunning{MaRF}
%
\author{Lorenzo Giusti\inst{1,2}  \and
Josue Garcia\inst{3} \and
Steven Cozine\inst{3} \\
\and
Darrick Suen\inst{3} \and
 Christina Nguyen\inst{3} 
 \and
Ryan Alimo\inst{2,4}
}
\authorrunning{L. Giusti et al.}
%
\institute{Sapienza University of Rome, Rome, Italy \and
Jet Propulsion Laboratory, California Institute of Technology, Pasadena, CA, 91109, USA \and
University of California, San Diego, San Diego, CA, USA \and OPAL AI INC., CA, USA}
\maketitle

\begin{abstract}
 The aim of this work is to introduce MaRF, a novel framework able to synthesize the Martian environment using several collections of images from rover cameras. The idea is to generate a 3D scene of Mars' surface to address key challenges in planetary surface exploration such as: planetary geology, simulated navigation and shape analysis. Although there exist different methods to enable a 3D reconstruction of Mars' surface, they rely on classical computer graphics techniques that incur high amounts of computational resources during the reconstruction process, and have limitations with generalizing reconstructions to unseen scenes and adapting to new images coming from rover cameras. The proposed framework solves the aforementioned limitations by exploiting Neural Radiance Fields (NeRFs), a method that synthesize complex scenes by optimizing a continuous volumetric scene function using a sparse set of images. To speed up the learning process, we replaced the sparse set of rover images with their neural graphics primitives (NGPs), a set of vectors of fixed length that are learned to preserve the information of the original images in a significantly smaller size. In the experimental section, we demonstrate the environments created from actual Mars datasets captured by Curiosity rover, Perseverance rover and Ingenuity helicopter, all of which are available on the Planetary Data System (PDS).
\end{abstract}

\section{Introduction}

\begin{figure}[t]
\centering
\includegraphics[width=.7\textwidth]{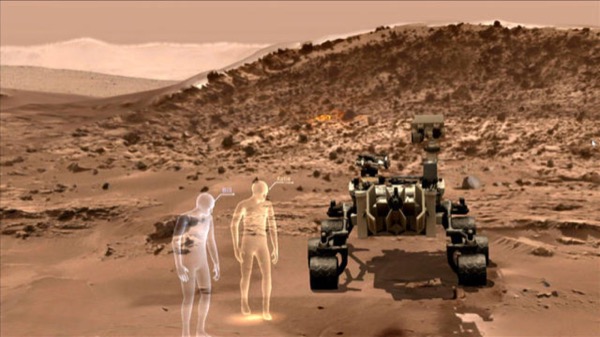}
\caption{Illustration of collaborative space exploration with scientists being virtually present as avatars onto the Mars' surface through mixed reality technology (Image credit: NASA/JPL-Caltech).}
\label{fig:onsight}
\end{figure}

A critical aspect of space exploration is understanding remote environments based on the limited data returned by spacecrafts. These environments are often too challenging or expensive to visit in person and can only be explored through images. While spacecraft imaging is an essential data source for learning about faraway locations, learning through two-dimensional images is drastically different from how geologists study Earth's environments \cite{abercrombie2019multi}. To address this challenge, mixed reality (MR), a human-computer interface, combines interactive computer graphics environments with real-world elements, recreating a whole sensory world entirely through computer-generated signals of an environment \cite{paelke2008integrating,mahmood2019improving}. MR overlays the computer-generated sensory signals onto the actual environment, enabling the user to experience a combination of virtual and real worlds concurrently. This technology allows scientists and engineers to interpret and visualize science data in new ways and experience environments otherwise impossible to explore. For instance, researchers at the NASA Goddard Space Flight Center (GSFC) have developed immersive tools for exploring regions from the depths of our oceans to distant stars and galaxies \cite{memarsadeghi2020virtual}. Another example is OnSight \cite{abercrombie2017onsight,beaton2020mission}, a framework developed through a collaboration between NASA JPL’s Operations Lab and Microsoft that creates an immersive 3-D terrain model of sites that Curiosity rover has visited and allows scientists to collaboratively study the geology of Mars as they virtually meet at those sites as shown in Fig. \ref{fig:onsight}. All these applications provide scientists and engineers an immersive experience of virtual presence in the field. Our contribution is an end-to-end framework that takes spacecrafts' image collections from JPL's Planetary Data System (PDS) \cite{mcmahon1996overview,padams2021fair}, filters them and learn the radiance field of Mars surface. Our framework is capable of recreating a synthetic view of the Martian scene in order to have a better sense of both the geometry and appearance of an environment that cannot be physically explored yet. Since the proposed mixed-reality framework captures both the scene's geometry and appearance information, enabling the synthesis of previously unknown viewpoints, it allows scientists and engineers to have a higher scientific return with respect to planetary geology, robotic navigation and many more space applications. Furthermore, the framework can also be used for compressing the images from the Mars Exploratory Rovers (MER) since the \emph{entire} scene is encoded in a Multi-Layer Perceptron (MLP) with a storage size of several orders of magnitude less than a single PDS image, all while preserving as much informative content as the original image collection.

This paper is structured as follows: Section \ref{sec:related_works} provides a review of the related works and technologies that use mixed reality for space applications, followed by the required mathematical background for understanding how a Neural Radiance Fields model for view synthesis is trained, and lastly the landscape of the most important techniques that exploit the original NeRF idea to solve the problems most closely related to ours. Section \ref{sec:marf} contains the core of this work. It provides a detailed description of the experimental setup and a review of the results obtained. Section \ref{sec:unc-estim} contains the methodology used to account for reconstruction instabilities using a bootstrap technique to measure the uncertainty of the reconstructions.

\section{Related Works}\label{sec:related_works}

In this section, we review mixed reality technologies for space applications, provide the required background for training a NeRF model, discuss related methods for photo-realistic view synthesis and explain the possibility of reducing the computational time required for achieving optimal results.

\subsection{Mixed Reality for Space Applications}\label{sec:mr-for-sa}

\begin{figure}[t]
    \centering
     \begin{subfigure}[b]{.4\textwidth}
         \centering
         \includegraphics[width=\textwidth]{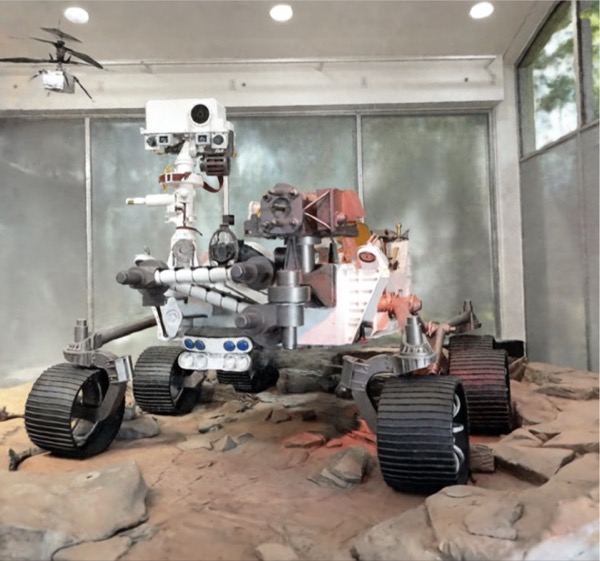}
         \caption{}
     \end{subfigure}
     \begin{subfigure}[b]{.4\textwidth}
         \centering
         \includegraphics[width=\textwidth]{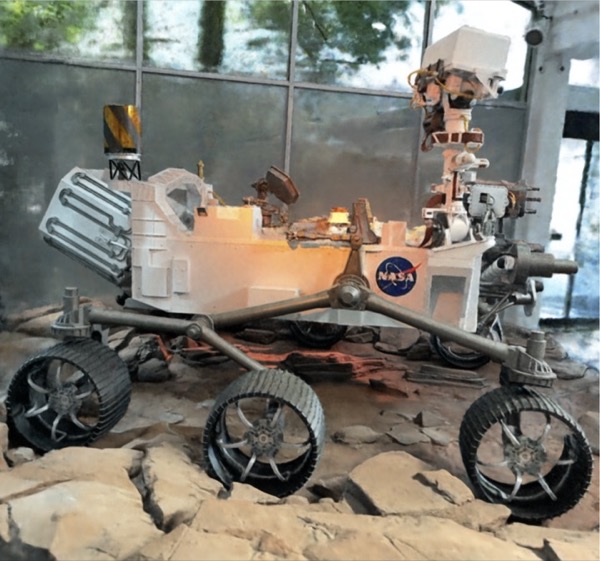}
         \caption{}
     \end{subfigure}
    \caption{Rendered view synthesis of the Perseverance rover at Jet Propulsion Laboratory: (a) frontal view; (b) side view.}
    \label{fig:rover-synth}
\end{figure}

Mixed reality has been successfully applied in many fields of \emph{aerospace engineering}. For example, it can be used to guide operators through a given task with the highest possible efficiency, achieved by superimposing 2-D and 3-D instructions and animations atop the working, real-world area  \cite{safi2019review}. Consider also that MR allows technicians to communicate with engineers while inspecting or repairing aircraft engines. Using this technology, a technician can be on any work site, wearing an immersive display with a camera, microphone, and speaker, while the engineer guides the technician remotely. For \emph{navigation and guidance} environments, immersive displays were initially used to provide pilots with flight information \cite{collinson2013introduction}. The use of mixed reality to display flight paths and obstacles can decrease pilot workload during landing operations in environments with scarcity of visibility. Augmented Reality (AR) systems have shown high potential for application in unmanned flight control and training. In \emph{simulations}, there have been numerous AR projects have investigate such technology and its application for pilot or astronaut training. For astronaut training and support, European Space Agency (ESA) developed the Computer Aided Medical Diagnostics and Surgery System (CADMASS) project \cite{nevatia2011computer}, which was designed to help astronauts perform mild to high-risk medical examinations and surgeries in space. The CADMASS project was designed to help, through mixed reality, those without medical expertise perform surgeries and similar procedures. 

\subsection{Neural Radiance Fields (NeRF)}\label{sec:nerf}

In this section, we provide a brief background for synthesizing novel views of large static scenes by representing the complex scenes as Neural Radiance Fields \cite{mildenhall2020nerf}. In NeRF, a scene is represented with a Multi-Layer Perceptron (MLP). Given the 3D spatial coordinate location $\mathbf{r} = (x, y, z)$ of a voxel and the viewing direction $d = (\theta, \phi)$ in a scene, the MLP predicts the volume density $\sigma$ and the view-dependent emitted radiance or color $\mathbf{c} = [R,G,B]$ at that spatial location for volumetric rendering. In the concrete implementation, $\mathbf{r}$ is first fed into the MLP, and outputs $\sigma$ and intermediate features, which are then fed into an additional fully connected layer to predict the color $c$. The volume density is hence decided by the spatial location $\mathbf{r}$, while the color $\mathbf{c}$ is decided by both $X$ and viewing direction $\mathbf{r}$. The network can thus yield different illumination effects for different viewing directions. This procedure can be formulated as:

\begin{equation}
    [R, G, B, \sigma] = F_\Theta(x,y,z,\theta,\phi),
\end{equation}

where $\Theta$ is the set of learnable parameters of the MLP. The network is trained by fitting the rendered (synthesized) views with the reference (ground-truth) views via the minimization of the total squared error between the rendered and true pixel colors in the different views. NeRF uses a rendering function based on classical volume rendering. Rendering a view from NeRF requires calculating the expected color $C(\mathbf{r})$ by integrating the accumulated transmittance $T$ along the camera ray $\mathbf{r}(t) = \mathbf{o}+t\mathbf{d}$, for $\mathbf{o}$ being the origin of cast ray from the desired virtual camera, with near and far bounds $t_n$ and $t_f$. This integral can be expressed as

\begin{equation}
    C(\mathbf{r}) = \int_{t_n}^{t_f} T(t) \sigma(\mathbf{r}(t)) c(\mathbf{r}(t), \mathbf{d}) dt,
\end{equation}\label{eq:exp_col}

where $T(t) = \exp{(-\int_{t_n}^t \sigma(\mathbf{r}(s)ds)}$ reflects the cumulative transmittance along the ray from $t_n$ to $t$, or the possibility that the beam will traverse the path without interacting with any other particle. Estimating this integral $C(\mathbf{r})$ for a camera ray traced across each pixel of the desired virtual camera is essential when producing a view from our continuous neural radiance field. To solve the equation \eqref{eq:exp_col}, the integral is approximated using the quadrature rule \cite{nelson95} by sampling a finite number of points $t_i$ uniformly along the ray $\mathbf{r}(t)$. Formally:

\begin{align*}
    \hat{C}(\mathbf{r}) &= \sum_{i=1}^{N} T_i (1- exp{(-\sigma_i ( \underbrace{t_{i+1} - t_{i}}_{\delta_i} ) )}) \mathbf{c}_i
\end{align*}
\label{eq:disc_exp_col}

where $\mathbf{c}_i$ is the color of the $i-th$ sample, $\sigma_i$ its density and $T_i= exp\left( -\sum_{j=1}^{i-1} \sigma_j \delta_j \right)$ estimates the transmittance. An example of NeRF results is shown in Fig. \ref{fig:rover-synth}.

\subsection{NeRF Landscape}\label{sec:nerf-landscape}

\begin{figure}[t]
    \centering
    \includegraphics[width=\textwidth]{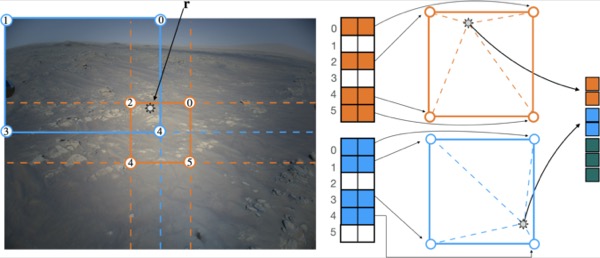}
    \caption{Example of multi-resolution hash encoding proposed in \cite{muller2022instant} for the 2D case on a picture collected from the Ingenuity helicopter. In this case, the number of resolution levels $L$ is set to be equal to $2$. The green boxes represent auxiliary inputs concatenated to the embedding vector before going into the neural network. }
    \label{fig:mash}
\end{figure}

Due to its computational time requirements, learning a view synthesis using Neural Radiance Fields might be intractable for huge natural scenes like the Martian environment. Despite the most successful breakthroughs in the field \cite{sitzmann2020implicit,zhang2020nerf++,yu2021pixelnerf,park2021nerfies}, the issue was partially solved with the introduction of Plenoxels \cite{yu21plenoxels}, where they proved how a simple grid structure with learned spherical harmonics coefficients may converge orders of magnitudes quicker than the fastest NeRF alternative without compromising the quality of the results. Recently, this structure was enhanced in \cite{muller2022instant}, in which they defined a multi-scale grid structure that looks up embeddings by querying coordinates in $\mathcal{O}(1)$ time using hash tables \cite{teschner2003optimized}. Essentially, their work focuses on the concept of leveraging a grid-like structure to store learning embeddings that may be interpolated to obtain correct values for each position. Multiple hash tables of differing resolutions are grouped in geometric progressions from the coarsest to the finer in the encoding structure. At each resolution level, they find the corners of the voxel containing the query coordinate, acquire their encodings, then interpolate them to create a collection of feature vectors for the provided coordinate, which is concatenated into a single input embedding; pictorially represented in Fig \ref{fig:mash}. There is a unique coordinate for each vertex in the coarse levels, but at the fine levels, the number of embeddings is less than the number of vertices, and a spatial hash function is used to extract the required feature vector. At further resolutions, it is feasible to trade off time requirements and quality of the results by increasing the size of the hash tables. These taught input embeddings allows to shrink the dimension of the network reducing drastically the overall computational time required for synthesize a scene, solving a critical bottleneck in past works. Implementing the entire process in a highly optimized low-level GPU code implementation, allows this approach to converge in a time that is several orders of magnitude less than the original NeRF implementation. Also, consider that the hash table may be queried in parallel on all resolution levels for all pixels, without requiring any form of control flow. Moreover, the cascading nature of the multi-resolution grid enables the use of a coarse one to one mapping between entries and spatial locations, and a fine “hash-table-like” structure with a spatial function that performs collision resolution via a gradient based method by averaging the colliding gradients, resulting in a sparse grid by design. If the distribution of points changes to a concentrated location anywhere in the picture, the collisions will become uncommon, and a better hashing function may be trained, thereby automatically adapting to the changing data on the fly without any task-specific structural adjustments.

\section{Mars Radiance Fields (MaRF)}\label{sec:marf}

\begin{figure}[t]
    \centering
    \includegraphics[width=\textwidth]{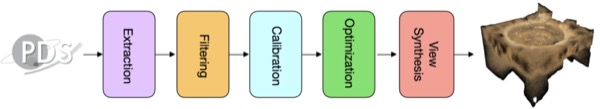}
    \caption{Sequence of operations required to get from the spacecrafts images to the neural view synthesis of the Mars' surface. The steps are, from left to right: (i) extract data from the Planetary Data System; (ii) filter image that do not meet the requirements needed for the reconstruction; (iii) calibrate the intrinsic and extrinsic camera parameters; (iv) training the algorithm to produce the view.}
    \label{fig:pipeline}
\end{figure}

In this section we provide the experimental setup used as well as the results obtained.

In particular we implemented a pipeline\footnote{Code available at: \url{https://github.com/lrnzgiusti/MaRF}} (Fig. \ref{fig:pipeline}) that is able to train a NeRF view synthesis from spacecraft images, available on the planetary data system (PDS), consisting of the following phases:

\textbf{Extraction}: To retrieve the spacecrafts' images, we queried the Planetary Data System by specifying the exact location of the images (i.e. \path{/mars2020/mars2020_mastcamz_ops_raw/}) and extracted \emph{all} the images present at that path across every \emph{sol}. Each image comes with an associated label files, which includes metrics such as the spacecraft's systems, the time of the image being taken and camera parameters (both intrinsics and extrinsics) in CAHVOR format \cite{tate2020mastcam}.

\begin{figure}[t]
    \centering
     \begin{subfigure}[b]{.4\textwidth}
         \centering
         \includegraphics[width=\textwidth]{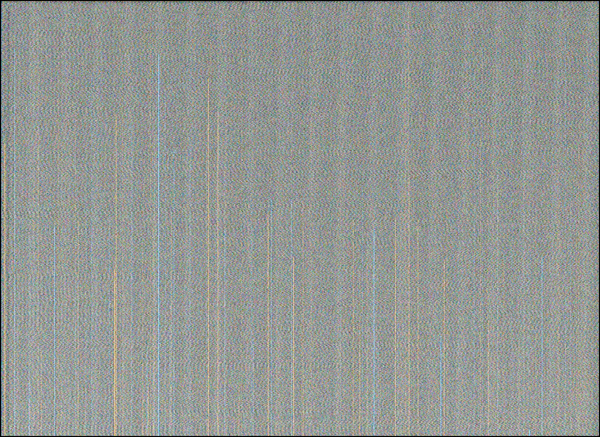}
         \caption{}
     \end{subfigure}
     \begin{subfigure}[b]{.4\textwidth}
         \centering
         \includegraphics[width=\textwidth]{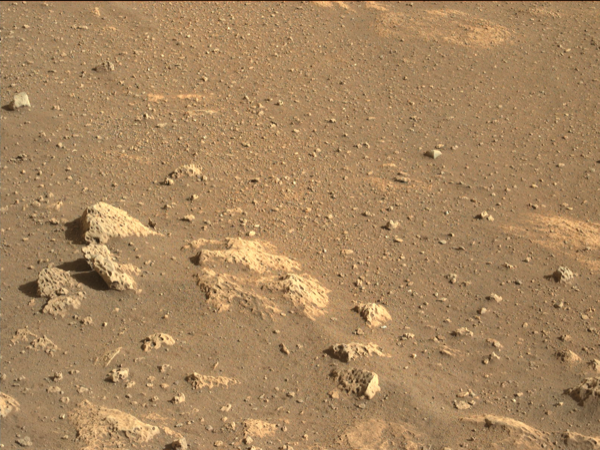}
         \caption{}
     \end{subfigure}
    \caption{Different images retrieved from the planetary data system (PDS). In the first case (a) the image is removed during the filtering stage while in (b) the preprocessing pipeline does not detect any defection and marks it as ready for the reconstruction.}
    \label{fig:good-bad}
\end{figure}

\textbf{Filtering}: Neural Radiance Fields provide a view synthesis' resolution proportional to the one of the images used for learning the scene. However, raw data provided by the \emph{extraction} phase might include images of varying resolution, including smaller thumbnail images; noisy images images and duplicates with different color filters applied. Thus, before training the view synthesis, the datasets are first forwarded through a bank of filters that keep images according to: (a) file dimension; (b) image shape; (c) duplicates up to a color filter; (d) grayscale; (e) color histogram; (f) blur detection. The first two filters (a-b) are parameterized using threshold values, i.e. if an image has a storage size less than a value $\beta$ or the shape is too small we remove since almost surely it does not meet the minimum quality requirements. The duplicate filter (c) utilizes a perceptual hash function \cite{zauner2010implementation} to identify photos with a near-identical hashes and remove images in which the environment is the same, but the color filters are different.
The grayscale filter (d) has two means of action: it identifies any photos with only one color channel, and filters those photos out. However, some grayscale photos have all three color channels. These photos can be identified because for each pixel, their RGB values will be equal.
The color histogram filter (e) analyzes the image channels in all photos that pass the previous filters and builds a histogram of the average color intensities. Then, a photo is filtered if more than half of the count of saturation values in a photo are more than a standard deviation away from the average saturation value. This methodology was adopted because Mars' images in general have consistent color channel values (Fig. \ref{fig:good-bad} (b)) while images that does not contain the Mars' environment will have far different color histograms (i.e. images containing parts of the spacecraft or glitched images Fig. \ref{fig:good-bad} (a)). 
The blur filter works by taking the variance of the Laplacian of the image, which numerically determines the sharpness of an image \cite{bansal2016blur}. By setting a threshold value $\tau$ on the sharpness we filter out all the images that have a sharpness value \emph{less} than $\tau$.

\textbf{Calibration}: For each image, the camera intrinsics and extrinsics models  are provided in CAHVOR format \cite{tate2020mastcam}. However, training a Neural Radiance Fields model requires these parameters to be expressed in a pinhole camera model \cite{hartley2003multiple}, which can be represented via the camera projection matrix as: 

\begin{equation}
    \mathbf{P} = \mathbf{K} \mathbf{R} \left[ \mathbf{I} \vert -\mathbf{t} \right]
\end{equation}

where $\mathbf{R}$ is a $3\times3$ rotation matrix,  $\mathbf{I}$ is a $3\times3$ identity matrix,  $\mathbf{t}$ is a $3\times1$ translation vector and $\mathbf{K}$ is the camera calibration matrix defined as:

\begin{equation}
    K = \begin{pmatrix}
            f_x & 0 & c_x\\
            0 & f_y & c_y \\
            0 & 0 & 1 \\
        \end{pmatrix}
\end{equation}

Here, $\mathbf{K}$  contains the intrinsic camera parameters where $f_x$ and $f_y$ are the focal lengths of the camera, which for square pixels assume the same value, and $c_x$ and $c_y$ are the offsets of the principal point from the top-left corner of the image. 

\begin{figure}[t]
\centering
\includegraphics[width=.55\textwidth]{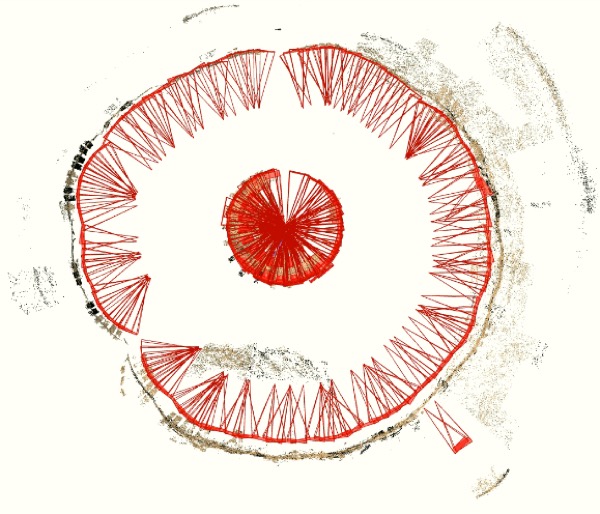}
\caption{Top-view of the camera calibration process for the Perseverance dataset.}
\label{fig:camera_calib}
\end{figure}

Although there exists techniques that provide a conversion between CAHVOR and pinhole camera models \cite{di2004cahvor} we rely the calibration phase on Structure from Motion (SfM) algorithms \cite{schoenberger2016sfm,schoenberger2016mvs} since the conversion defined in \cite{di2004cahvor} requires parameters not readily available on the PDS. An illustration of the results from the calibration process is shown in Fig. \ref{fig:camera_calib}.

\textbf{Optimization}:  To properly tune the model's hyper-parameters, we rely on random search \cite{bergstra2012random}. The metric used to quantitatively measure the reconstruction quality is the \emph{peak signal-to-noise ratio} between the underlying scene represented by $\mathcal{S}$ described by a set of images  $\{I_{j} \}_{j=1}^N$, where each image $I_j$ is captured by a viewpoint $\mathbf{r}_j$  and the view synthesis $\hat{\mathcal{S}}$, composed by a set of rendered images $\{\hat{I}_{j} \}_{j=1}^N$ where the elements $\hat{I}_j$ are rendered from the same viewpoint of the ground truth $\mathbf{r}_j$. Then, for a single image, the $\mathit{PSNR}$ between the ground truth and the synthesized one is computed as:

\begin{align}
\mathit{PSNR(I,\hat{I})} &= 10\cdot \log_{10}\left( \frac{ \mathit{max}(I^{2})}{\mathit{MSE(I, \hat{I})}} \right),
\end{align}

where $\mathit{MSE(I, \hat{I})} =  \mathit{ \frac{ \left( I-\hat{I}\right)^{2}}{W\cdot H \cdot C}}$. Then, to get the peak signal-to-noise ratio between $\mathcal{S}$ and $\hat{\mathcal{S}}$, we compute the average on the peak signal-to-noise ratios of the images that compose the scene.

\begin{figure}[t]
     \centering
     \begin{subfigure}[b]{0.3\textwidth}
         \centering
         \includegraphics[width=\textwidth]{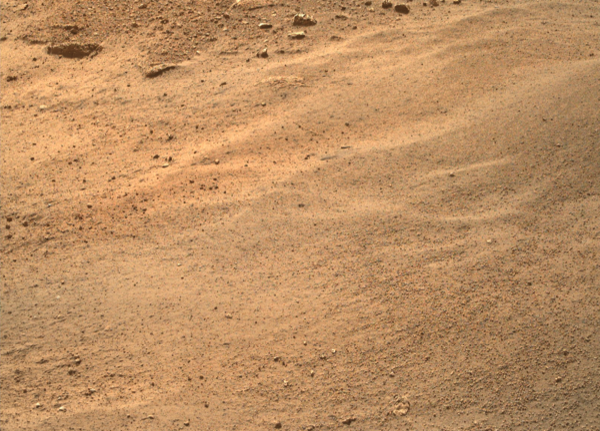}
         \caption{}
         \label{fig:pers-base}
     \end{subfigure}
     \hfill
     \begin{subfigure}[b]{0.3\textwidth}
         \centering
         \includegraphics[width=\textwidth]{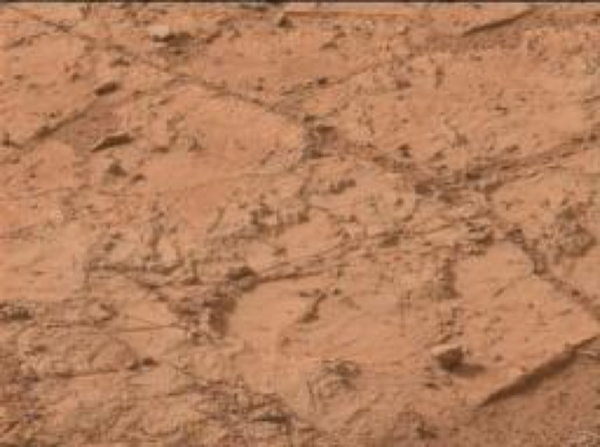}
         \caption{}
         \label{fig:msl-base}
     \end{subfigure}
     \hfill
     \begin{subfigure}[b]{0.3\textwidth}
         \centering
         \includegraphics[width=\textwidth]{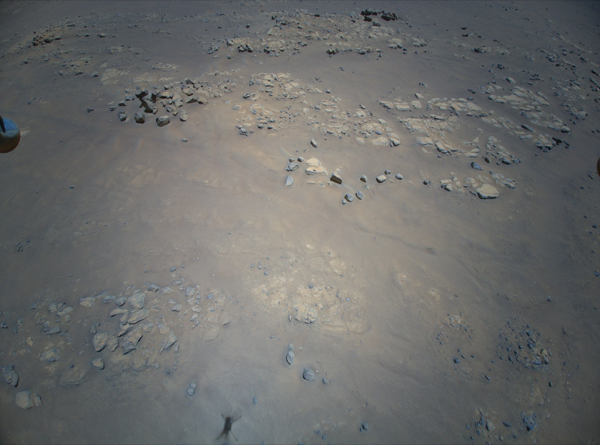}
         \caption{}
         \label{fig:heli-base}
     \end{subfigure}
     \begin{subfigure}[b]{0.3\textwidth}
         \centering
         \includegraphics[width=\textwidth]{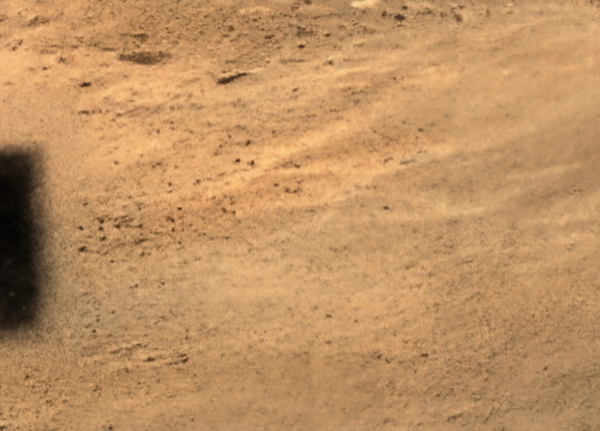}
         \caption{}
         \label{fig:pers-recon}
     \end{subfigure}
     \hfill
     \begin{subfigure}[b]{0.3\textwidth}
         \centering
         \includegraphics[width=\textwidth]{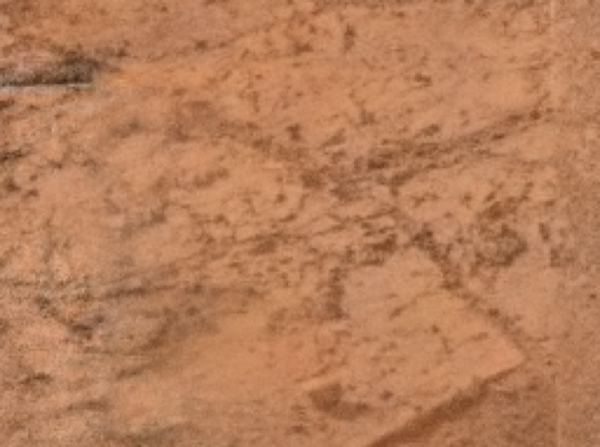}
         \caption{}
         \label{fig:msl-recon}
     \end{subfigure}
     \hfill
     \begin{subfigure}[b]{0.3\textwidth}
         \centering
         \includegraphics[width=\textwidth]{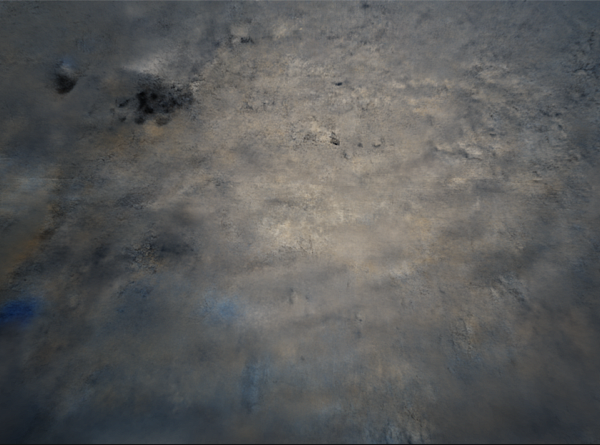}
         \caption{}
         \label{fig:heli-recon}
     \end{subfigure}
        \caption{A demonstration of the view synthesis quality for spacecrafts' images. Top row: Reference test images of \textbf{(a)} Perseverance;  \textbf{(b)} Curiosity;  \textbf{(c)} Ingenuity datasets for which we evaluated the proposed framework. Bottom row: rendered view synthesis of the reference images using MaRF. The rendering of the reconstruction for the Perseverance dataset \textbf{(d)} contains a black rectangle due to a misalignment during the camera calibration.    }
        \label{fig:grid-results}
\end{figure}

\textbf{View Synthesis}: For the scope of this work, we used different datasets coming from the following sources: \textbf{(a)} Curiosity rover's science cameras \cite{bell2017mars}; \textbf{(b)} Ingenuity helicopter's color camera \cite{balaram2021ingenuity}; \textbf{(c)} Perseverance rover's Mastcam-Z \cite{bell2021mars}. In Table \ref{tab:multicol}, we shown a quantitative assessment of the synthesis process by comparing the peak signal to noise ratio (PSNR) of the proposed framework for the aforementioned datasets and listing the results of our method after training for 10 seconds to 15 minutes. For the Perseverance dataset, the PSNR decreases with respect to the training time. This fact is due to a slight misalignment of the camera models involved during the training process leading to sparse black spots in the view synthesis as shown in Fig. \ref{fig:pers-recon}. Although this misalignment influences the quantitative results of the scene, the overall quality of the reconstruction is not compromised. 

\begin{table}[!htb]
\begin{center}
\begin{tabular}{lccc}
\hline
\multicolumn{4}{Sc}{\textbf{Quantitative Results}}\\
\toprule
\multicolumn{1}{c}{}  & \multicolumn{1}{c}{Curiosity} & \multicolumn{1}{c}{Perseverance} & \multicolumn{1}{c}{Ingenuity}  \\ \hline
{\textit{PSNR} \textbf{(10s)}}   & 
\multicolumn{1}{c}{$23.02$} & 
\multicolumn{1}{c}{$18.09$} & 
\multicolumn{1}{c}{$22.07$}\\
{\textit{PSNR} \textbf{(60s)}}       & 
\multicolumn{1}{c}{$26.22$} & 
\multicolumn{1}{c}{$18.08$} & 
\multicolumn{1}{c}{$23.00$}\\
{\textit{PSNR} \textbf{(5m)}}            & 
\multicolumn{1}{c}{$31.01$} & 
\multicolumn{1}{c}{$17.82$} & 
\multicolumn{1}{c}{$23.86$}\\
{\textit{PSNR} \textbf{(10m)}}            & 
\multicolumn{1}{c}{$34.71$} & 
\multicolumn{1}{c}{$17.71$} & 
\multicolumn{1}{c}{$24.25$}\\
{\textit{PSNR} \textbf{(15m)}} & 
\multicolumn{1}{c}{$36.82$} & 
\multicolumn{1}{c}{$17.63$} & 
\multicolumn{1}{c}{$24.24$}\\
\bottomrule
\end{tabular}
\end{center}
\caption{Peak signal to noise ratio (PSNR) of MaRF across the selected spacecraft datasets for different execution times of the reconstruction. }
\label{tab:multicol}
\end{table}

\section{Bootstrapping the Uncertainty}\label{sec:unc-estim}

\begin{figure}[t]
    \centering
    \begin{subfigure}[b]{.3\textwidth}
         \centering
         \includegraphics[width=\textwidth]{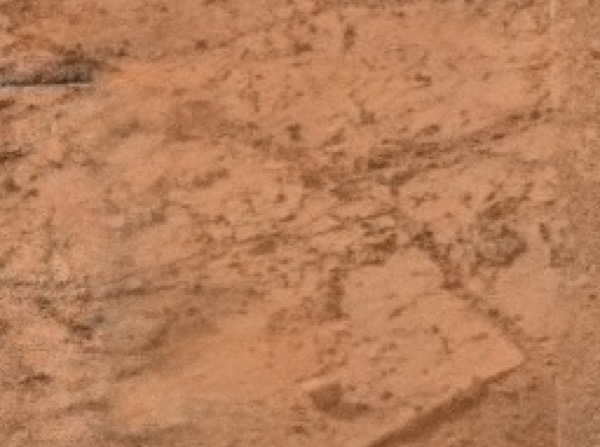}
         \caption{}
     \end{subfigure}
     \hfill
     \begin{subfigure}[b]{.3\textwidth}
         \centering
         \includegraphics[width=\textwidth]{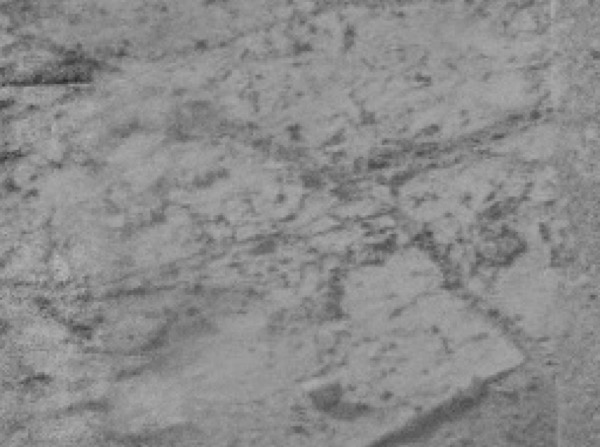}
         \caption{}
     \end{subfigure}
     \hfill
     \begin{subfigure}[b]{.3\textwidth}
         \centering
         \includegraphics[width=\textwidth]{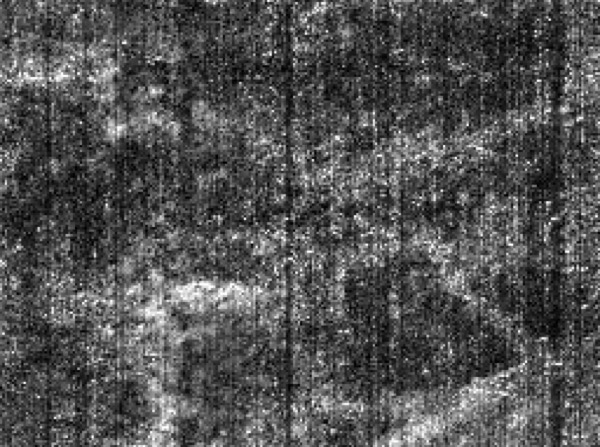}
         \caption{}
     \end{subfigure}
    \caption{Uncertainty estimation. \textbf{(a)} Pixel-wise mean out of the bootstrapped renderings. \textbf{(b)} Grayscale version of \textbf{(a)}, used to compute the uncertainty maps. \textbf{(c)} Pixel-wise uncertainty map; here uncertainty is proportional to the pixel intensity.  }
    \label{fig:unc}
\end{figure}

In this section we propose a technique to account for fluctuations during the training of the Mars' radiance field, as well as computing a view synthesis' confidence in terms of an uncertainty map of the reconstruction. The aim of these techniques is to provide useful insights and allow scientist and engineering teams to stay in the loop when it comes to decision making for the Mars' missions, maximizing the scientific return. For example, in a simulated environment, the uncertainty map provides a support to the navigation team to determine the best sequencing plan, while being aware of potential pitfalls. To compute the uncertainty map of the synthesized scene, we follow the bootstrap method  \cite{efron1994introduction}. In particular, once the model's hyper-parameters have been optimized, we repeat the training process $B$ times using the same setup and then the spatial locations in which the model is unstable is computed as the pixel-wise standard deviation of the rendered frames. Formally:

\begin{enumerate}
    \item Optimize the models' hyper-parameters according to the method defined in the section \ref{sec:marf}.
    \item Let $\{\hat{\mathcal{S}}_b\}_{b=1}^B$ be the set of bootstrapped view synthesis of the same scene.
    \item Render each scene to produce a multi-set $\{\{\hat{I}_{j,b}\}_{j=1}^N\}_{b=1}^B$ of bootstrapped rendered images from the viewpoints $\mathbf{r}_j$.
    \item Let  $\tilde{I}_{j,b}$ be the gray-scale version of $\hat{I}_{j,b}$ and stack them into $\mathbf{\tilde{I}}_j$, a 3D tensor in which the bootstrap results are placed along the third dimension.
    \item Let $\mathbf{\tilde{I}}_{j} = \frac{1}{B}\sum_b  \mathbf{\tilde{I}}_{j,b}$ be the expected rendering at viewpoint $\mathbf{r}_j$.
    \item Let $\Sigma_j = \sum_b \sigma\left( \mathbf{\tilde{I}}_{j,b} \right)$  be the uncertainty map at viewpoint $\mathbf{r}_j$, computed as the pixel-wise standard deviation along its third dimension.
\end{enumerate}

By this point, we have the uncertainty maps for all the viewpoints of the scene. With this, we are able to put them in a sequence to provide a fly-through video of the geometric and texture uncertainties. 

\section{Conclusion and Discussion}\label{sec:conclusions}

In this work we presented \emph{MaRF}, an end-to-end framework able to convert spacecrafts images available on the Planetary Data System (PDS) into an synthetic view of the Mars surface by exploiting Neural Radiance Fields. It consists of an extract, transform, load (ETL) pipeline that (i) retrieves images from the PDS; (ii) filters them according to a minimum quality requirements; (iii) calibrates the cameras using structure from motion (SfM) algorithms, (iv) Optimizes the model's hyper-parameters and (v) obtains the view synthesis of the Martian environment by learning its radiance field using a neural network that optimizes a continuous volumetric scene function using the sparse set of images that have been calibrated during phase (iii). We have evaluated the proposed technique on three different spacecraft datasets: Curiosity rover, Perseverance rover and Ingenuity helicopter. We also equip our framework with a method that is able to account for reconstruction instabilities and provide an estimation of the uncertainty of the reconstruction. We believe that equipping future space missions with tools like MaRF, at the intersection of mixed reality and artificial intelligence, will make a significant impact in terms of scientific returns, allowing collaborative space exploration, enhancing planetary geology and having a better sense of remote environments that cannot yet be explored.


\clearpage
%
%
\bibliographystyle{splncs04}
\bibliography{refs}
\end{document}